\documentclass[pdflatex,sn-mathphys-num]{sn-jnl}

\usepackage{graphicx}%
\usepackage{multirow}%
\usepackage{amsmath,amssymb,amsfonts}%
\usepackage{amsthm}%
\usepackage{mathrsfs}%
\usepackage[title]{appendix}%
\usepackage{xcolor}%
\usepackage{textcomp}%
\usepackage{manyfoot}%
\usepackage{booktabs}%
\usepackage{algorithm}%
\usepackage{algorithmicx}%
\usepackage{algpseudocode}%
\usepackage{listings}%
\usepackage{mlmath}
\usepackage{longtable} 

\theoremstyle{thmstyleone}%
\theoremstyle{thmstyletwo}%

\theoremstyle{thmstylethree}%
\definecolor{headerred}{RGB}{220,64,28}
\definecolor{rowpink}{RGB}{242,220,220}
\definecolor{gainlow}{RGB}{190,50,40}     
\definecolor{gainhigh}{RGB}{0,120,140}    
\begin{document}

\title[Small Initialization Matters for Large Language Models]{Small Initialization Matters for Large Language Models}


\author[1]{\fnm{Liangkai} \sur{Hang}}\email{hangliangkai@sjtu.edu.cn}
\equalcont{These authors contributed equally to this work.}

\author[1]{\fnm{Junjie} \sur{Yao}}\email{yaojj22@sjtu.edu.cn}
\equalcont{These authors contributed equally to this work.}

\author[3,4]{\fnm{Zhiyu} \sur{Li}}
\author[3,4]{\fnm{Feiyu} \sur{Xiong}}
\author[3,4]{\fnm{Hongkang} \sur{Yang}}

\author*[1,2]{\fnm{Zhi-Qin John} \sur{Xu}}\email{xuzhiqin@sjtu.edu.cn}

\affil[1]{\orgdiv{School of Mathematical Sciences}, \orgname{Shanghai Jiao Tong University}, \orgaddress{ \city{Shanghai}, \postcode{200240}, \country{China}}}

\affil[2]{\orgdiv{Institute of Natural Sciences}, \orgname{Shanghai Jiao Tong University}, \orgaddress{ \city{Shanghai}, \postcode{200240}, \country{China}}}

\affil[3]{\orgname{MemTensor (Shanghai) Technology Co., Ltd.}}

\affil[4]{\orgname{Institute for Advanced Algorithms Research},  \orgaddress{ \city{Shanghai}, \country{China}}}


\abstract{
Large language models provide a tractable system for asking how intelligence itself emerges, rather than only how LLMs can be engineered. Although progress is usually attributed to scale, data and architecture, we show that parameter initialization is a gene-like determinant of training and, in particular, of model capacity. Reducing the initialization scale consistently improves pretraining, with the largest gains on reasoning-demanding tasks. We identify two widely used empirical settings that restrain the advantage of small initialization, and show how relaxing them restores favorable scaling. We further uncover a critical initialization that balances the reasoning and training. Mechanistically, small initialization drives a distinct developmental trajectory: parameters first condense into low-complexity structures and later expand into richer representations, giving concrete form to the idea that compression is intelligence. Token-level analyses show that the gains concentrate on non-trivial, context-constrained predictions rather than all tokens uniformly. These results motivate a simple $\gamma$-initialization rule: expose initialization rage as an explicit knob and use small initialization by default, an almost cost-free intervention that improves pretraining and strengthens reasoning across model scales.}

\maketitle

\section{Introduction}
Beyond their practical role as engineered systems, large language models (LLMs) offer an experimental window into how intelligence can emerge from scale, data, optimization and architecture. Their recent advances have largely come from increasing scale~\citep{brown2020language,kaplan2020scaling}, refining data and optimization~\citep{NEURIPS2022_c1e2faff,hu2022lora,liu2024sophia,guo2025deepseek}, or modifying architectures~\citep{vaswani2017attention,devlin2019bert,fedus2022switch,gu2023mamba}. Yet parameter initialization remains a critical design choice underpinning modern deep learning~\citep{lecun2015deep}. Once considered settled by handcrafted heuristics such as Xavier~\citep{glorot2010understanding}, LeCun~\citep{lecun2012efficient} and Kaiming~\citep{he2015delving} initialization, it has resurfaced in the large-model era, where a tailored scheme now demands costly trial and error. In this work, we demonstrate that the scale of parameter initialization profoundly influences the learning process of LLMs, as a concrete manifestation to support the viewpoint that compression essentially embodies intelligence.

Large initialization makes networks behave like kernel methods~\citep{jacot2018neural,chizat2019lazy,woodworth2020kernel}, whereas small initialization drives them into a nonlinear feature-learning, or condensed, regime, in which the weight vectors within a layer first align along a few shared directions and later develop richer structure~\citep{luo2021phase,zhou2022towards,kunin2024get}. 
Small initialization has been shown to bias models toward reasoning and improve generalization, though largely on simplified architectures and synthetic tasks~\citep{zhang2024initialization,yao2025analysis,zhang2025complexity}. Whether these benefits carry over to realistic LLM pretraining, and persist as models scale up, remains largely unexplored. Here we systematically examine whether, when, and why small initialization improves LLM performance. 

We parameterize each weight matrix as $W_{ij}\sim\mathcal{N}(0,\sigma^2),\sigma=d_{\mathrm{in}}^{-\gamma}$, where the initialization rate $\gamma=1/2$ recovers the standard scale such as Xavier-like scale~\cite{glorot2010understanding} and larger $\gamma$ yields smaller initialization. Training LLMs across a range of $\gamma$, we find that simply reducing the initialization scale lowers the pretraining loss, though this advantage fades as models grow. 
This weakening is not an intrinsic limitation of small initialization but a consequence of specific architectural components: layer normalization~\cite{ba2016layer} obscures differences in scale through its constant $\epsilon$, while small initialization simultaneously intensifies the attention sink~\cite{xiao2024streamingllm}. Reducing $\epsilon$ and introducing gated attention \citep{qiu2026gated} unlock its latent benefit, yielding a markedly better scaling law with model size. 

Our results further indicate that an equal balance between the identity and residual pathways—achieved at $\gamma = 1$—yields the best performance. Finally, we find that small-initialization models follow a low-to-high complexity trajectory: their weight matrices first condense into low-dimensional structures and later expand into a richer space, a condensation phenomenon \citep{xu2025overview} shared by models from multilayer perceptrons to Transformers in condensed regimes. A token-level analysis of the loss further shows that the gains are not spread evenly across tokens but concentrate on a subset of non-trivial ones. 



Together, these results establish the scale of initialization as both a practical and a mechanistic design axis for LLMs: small initialization substantially improves pretraining, $\gamma=1$ marks a critical balance point, and the resulting training dynamics follow a clear condensed pattern. More broadly, we argue for treating $\gamma$ as an explicit initialization parameter, and for adopting $\gamma$-initialization as a built-in initializer in mainstream deep learning frameworks, with $\gamma=1$ as the default.

\section{Result}

\subsection{Why small initialization fails to scale, and how to fix it}
\begin{figure}[htbp]
    \centering
    \includegraphics[width=1\linewidth]{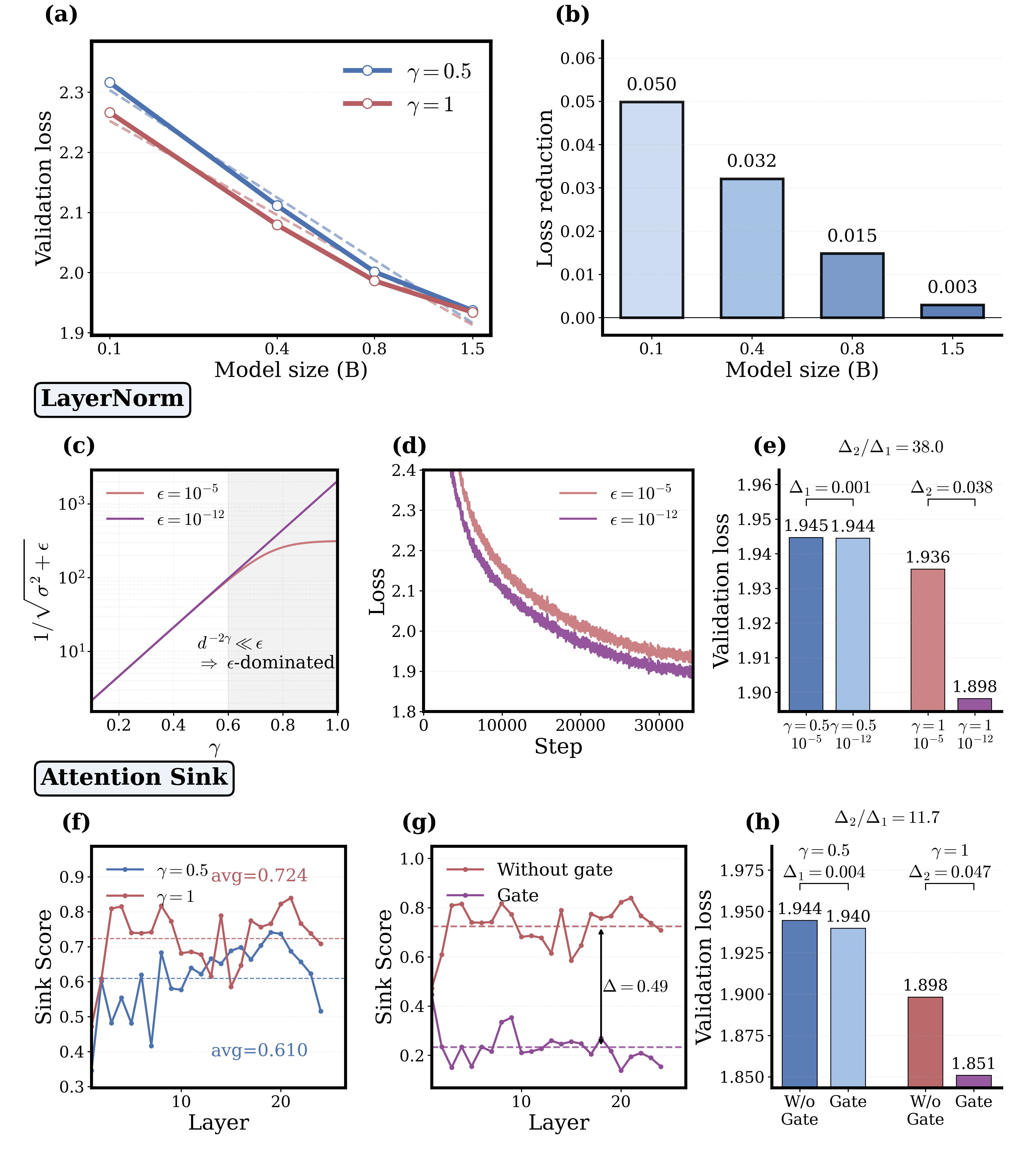}
    \caption{(a) Validation loss across different model sizes under two initialization scales, $\gamma=0.5$ and $\gamma=1$. (b) Absolute validation loss reduction between $\gamma=0.5$ and $\gamma=1$ across model scales. (c) Effective RMSNorm scaling factor as a function of $\gamma$ under two different $\epsilon$ settings, where $d=2048$. (d) Training loss curves for 1.5B models trained under $\gamma=1$ with different $\epsilon$ values. (e) Final validation loss comparison under different combinations of initialization scale $\gamma$ and RMSNorm $\epsilon$. (f) Layer-wise attention sink scores under different initialization scales in 1.5B models. (g) Layer-wise attention sink scores with and without gated attention under $\gamma=1$. (h) Final validation loss comparison for models trained with and without gated attention under different initialization scales.}
    \label{fig:small_init_gain}
\end{figure}

We first examine whether reducing the initialization scale benefits LLM pretraining, training models of different sizes under the standard scale $\gamma=0.5$ and a smaller scale $\gamma=1$. Small initialization consistently lowers the validation loss across all model scales (Figure~\ref{fig:small_init_gain}a,b). The gain, however, shrinks with size: moving from $\gamma=0.5$ to $\gamma=1$ reduces the loss by about $0.05$ for the 0.1B model but only $0.003$ for the 1.5B model. This raises a central question: why does the benefit of small initialization vanish at scale?

We show that this shrinkage is not a failure of small initialization but a consequence of architectural components that suppress its effect. We identify two. First, layer normalization enters an $\epsilon$-dominated regime once the hidden-state variance becomes small, masking the scale difference. Second, small initialization aggravates the attention sink, concentrating attention on the first token. Removing both barriers restores the gain, as we show below.

\paragraph{Adjusting the RMSNorm constant}
RMSNorm rescales inputs $\vh$ by a factor proportional to $\left(\sigma^2(\vh)+\varepsilon\right)^{-1/2}$, where $\sigma^2(\vh) = 1/d\sum_{i}^d\vh^2_i$ and $\varepsilon$ is the stability constant. Small initialization shrinks $\sigma^2(\vh)$, and once $\sigma^2(\vh)\lesssim\varepsilon$ the factor is governed by $\varepsilon$ rather than the $\vh$; further reducing the scale then no longer changes the normalization, masking the effect of small initialization. Taking $\vh_i\sim\mathcal{N}(0,d^{-2\gamma})$ so that $\sigma^2(\vh)=d^{-2\gamma}$. Consider with $d=2048$, then $\sigma^2(\vh)=4.8\times10^{-4}$ when $\gamma=0.5$ and $2.4\times10^{-7}$ when $\gamma=1$, which is much smaller that the common $\varepsilon=10^{-5}$. This indicates that the common value $\varepsilon=10^{-5}$ saturates the factor for $\gamma>0.6$, whereas $\varepsilon=10^{-12}$ keeps it sensitive to $\gamma$ over a far wider range (Figure~\ref{fig:small_init_gain}c).

The prediction holds in training. For 1.5B models at $\gamma=1$, lowering $\varepsilon$ from $10^{-5}$ to $10^{-12}$ markedly decreases the loss (Figure~\ref{fig:small_init_gain}d, $\sim0.038$) while at $\gamma=0.5$ the same reduction barely helps (Figure~\ref{fig:small_init_gain}e, $\sim 0.001$).

\paragraph{Mitigating the attention sink}
The second barrier is the attention sink, the tendency of LLMs to place disproportionate attention on the first token~\cite{xiao2024streamingllm,barbero2025why,ICLR2025_f1b04fac,Yu2024unveiling}. Measuring the sink score (see~\eqref{eq:sink_score}), the average attention mass on the first token, we find that small initialization strengthens it: the per-layer average under $\gamma=1$ exceeds that under $\gamma=0.5$ by about $0.11$ in 1.5B models (Figure~\ref{fig:small_init_gain}f). 

To remove it, we apply gated attention, which gates the output of each attention head~\cite{qiu2026gated}. Gating sharply reduces the sink score under $\gamma=1$ (Figure~\ref{fig:small_init_gain}g), and this translates into performance: at $\gamma=0.5$ gating barely lowers the loss ($\sim0.004$), whereas at $\gamma=1$ it produces a much larger decrease (Figure~\ref{fig:small_init_gain}h, $\sim0.047$). Mitigating the attention sink therefore substantially strengthens the effect of small initialization. It's noted this ablation experiment is implemented with $\varepsilon=10^{-12}$.

\subsection{Unleashing the benefit of small initialization}
Once the two architectural barriers are removed, the benefit of small initialization emerges in full. Combining the adjustments, reducing $\varepsilon$ to $10^{-12}$ and adding gated attention, leaves the loss essentially unchanged at $\gamma=0.5$ but markedly amplifies the gain at $\gamma=1$, and this gain persists as model size grows (Figure~\ref{fig:scalinglaw}a). The comparison indicates that small initialization can improve the effective model size by approximately $79\%$. The adjustments thus matter only under small initialization: they do not improve training on their own, but release a benefit of small initialization that is otherwise hidden.

This advantage extends to downstream capability. On 1.5B models with the adjustments, $\gamma=1$ outperforms $\gamma=0.5$ on benchmarks spanning knowledge, commonsense reasoning, and math (Table~\ref{tab:eval}), with absolute gains exceeding $4\%$ on TriviaQA, HellaSwag, GSM8K, and MATH500. Small initialization, properly supported by the architecture, therefore yields both lower loss and stronger task performance.

\begin{table}[t]
\centering
\small
\setlength{\tabcolsep}{8pt}
\renewcommand{\arraystretch}{1.15}
\begin{tabular}{llccc}
\toprule
Category & Benchmark & $\gamma = 0.5$ & $\gamma = 1$ & Gain \\
\midrule
\multirow{4}{*}{Knowledge}
& ARC-C~\cite{clark2018think}    & 27.28 & 29.01 & \textcolor{gainlow}{+1.73} \\
& TriviaQA~\cite{2017arXivtriviaqa} &  8.80 & 13.00 & \textcolor{gainhigh}{+4.20} \\
& MMLU~\cite{hendrycks2021ethics,hendryckstest2021}     & 25.32 & 27.11 & \textcolor{gainlow}{+1.79} \\
\midrule
\multirow{3}{*}{Commonsense Reasoning}
& HellaSwag~\cite{zellers2019hellaswag} & 38.74 & 43.38 & \textcolor{gainhigh}{+4.64} \\
& BBH~\cite{suzgun2023challenging}       & 28.18 & 29.50 & \textcolor{gainlow}{+1.32} \\
& SocialIQA~\cite{sap2019social} & 40.17 & 40.22 & \textcolor{gainlow}{+0.05} \\
\midrule
\multirow{2}{*}{Math}
& GSM8K~\cite{cobbe2021training}   &  2.70 &  7.60 & \textcolor{gainhigh}{+4.90} \\
& MATH500~\cite{hendrycks2021measuring} &  6.00 & 10.00 & \textcolor{gainhigh}{+4.00} \\
\midrule
\multicolumn{2}{l}{Average} & 22.47 & 24.98 & \textbf{\textcolor{gainlow}{+2.51}} \\
\bottomrule
\end{tabular}
\caption{Evaluation results of the 1.5B models under standard initialization ($\gamma=0.5$) and small initialization ($\gamma=1$). The gain column reports the absolute improvement of $\gamma=1$ over $\gamma=0.5$. Gains smaller than 4 are shown in red, while gains greater than or equal to 4 are shown in green.}
\label{tab:eval}
\end{table}

\subsection{Extension to mixture-of-experts models}

\begin{figure}[htbp]
    \centering
    \includegraphics[width=1\linewidth]{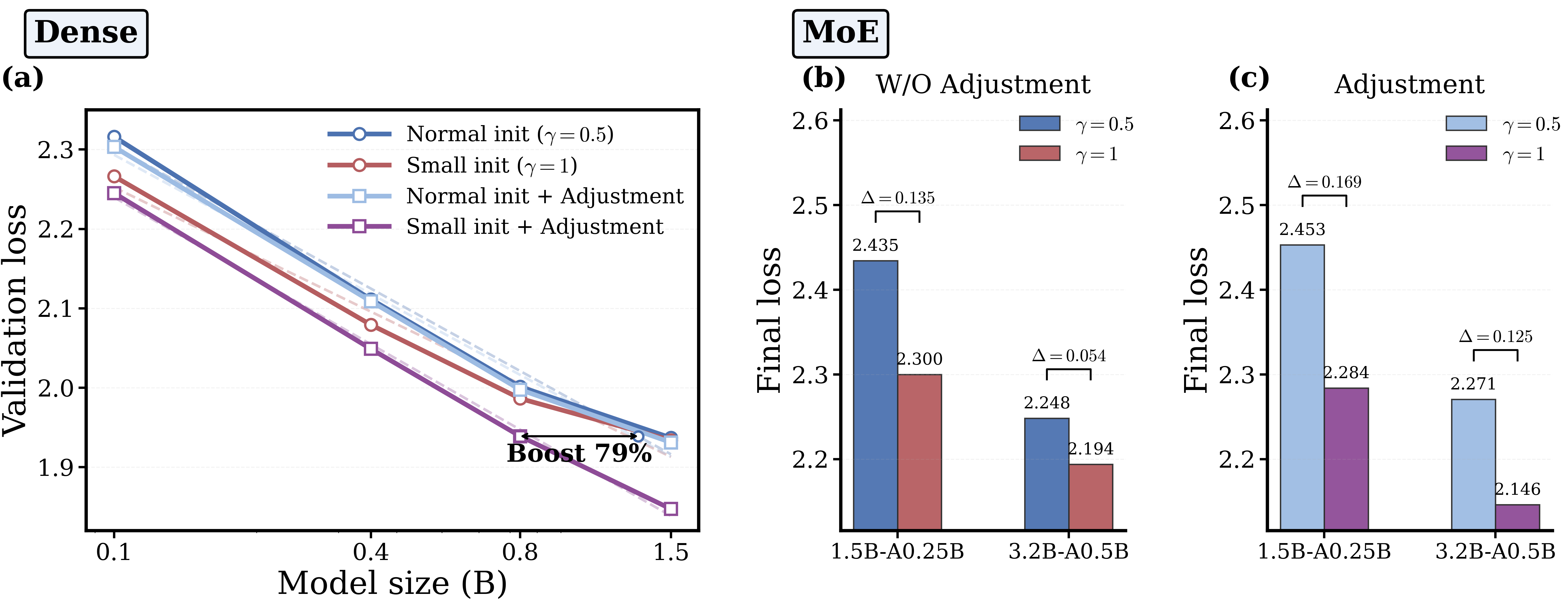}
    \caption{(a) Validation loss across different model sizes under $\gamma=0.5$ and $\gamma=1$, with and without the architectural adjustments. (b) Validation loss of MoE models without the adjustments under $\gamma=0.5$ and $\gamma=1$. (c) Validation loss of MoE models with the adjustments under $\gamma=0.5$ and $\gamma=1$.}
    \label{fig:scalinglaw}
\end{figure}

To test whether the effect is specific to dense models, we repeat the experiments on mixture-of-experts (MoE) models in two configurations: 1.5B total (0.25B active) and 3B total (0.5B active) parameters, each trained at $\gamma=0.5$ and $\gamma=1$. The MoE results mirror the dense case (Figure~\ref{fig:scalinglaw}b,c): small initialization clearly lowers the loss at the smaller size, the gain weakens at the larger size under the standard architecture, and reducing $\varepsilon$ to $10^{-12}$ together with gated attention recovers and amplifies it. Both the limitation and its remedy therefore carry over, establishing small initialization as a broadly useful strategy across architectures. 
\subsection{How small should initialization be?}

Given these gains, should the scale be pushed as small as possible, that is, $\gamma$ made arbitrarily large? We find that it should not, for a reason rooted in the residual structure of a Transformer. Each layer adds a residual update to an identity pathway; when the weights are too small, these updates vanish and the identity pathway dominates, leaving the network unable to transform its input during early training. A useful initialization must keep the residual updates comparable to the identity pathway.

This balance can be made precise. For a pre-norm Transformer of $L$ layers with final hidden state $\vh_L$ and embedded input $\ve$, we define the residual flow as $\vh_L-\ve$. Under small initialization, its relative scale satisfies $\|\vh_L-\ve\|_2/\|\ve\|_2\asymp d^{1-\gamma}$ (derivation in Appendix~\ref{app:proof}). The residual flow thus dominates the embedding for $\gamma<1$, matches it at $\gamma=1$, and becomes negligible for $\gamma>1$, where the network is initialized close to an identity mapping.

\begin{figure}[htpb]
    \centering
    \includegraphics[width=0.9\linewidth]{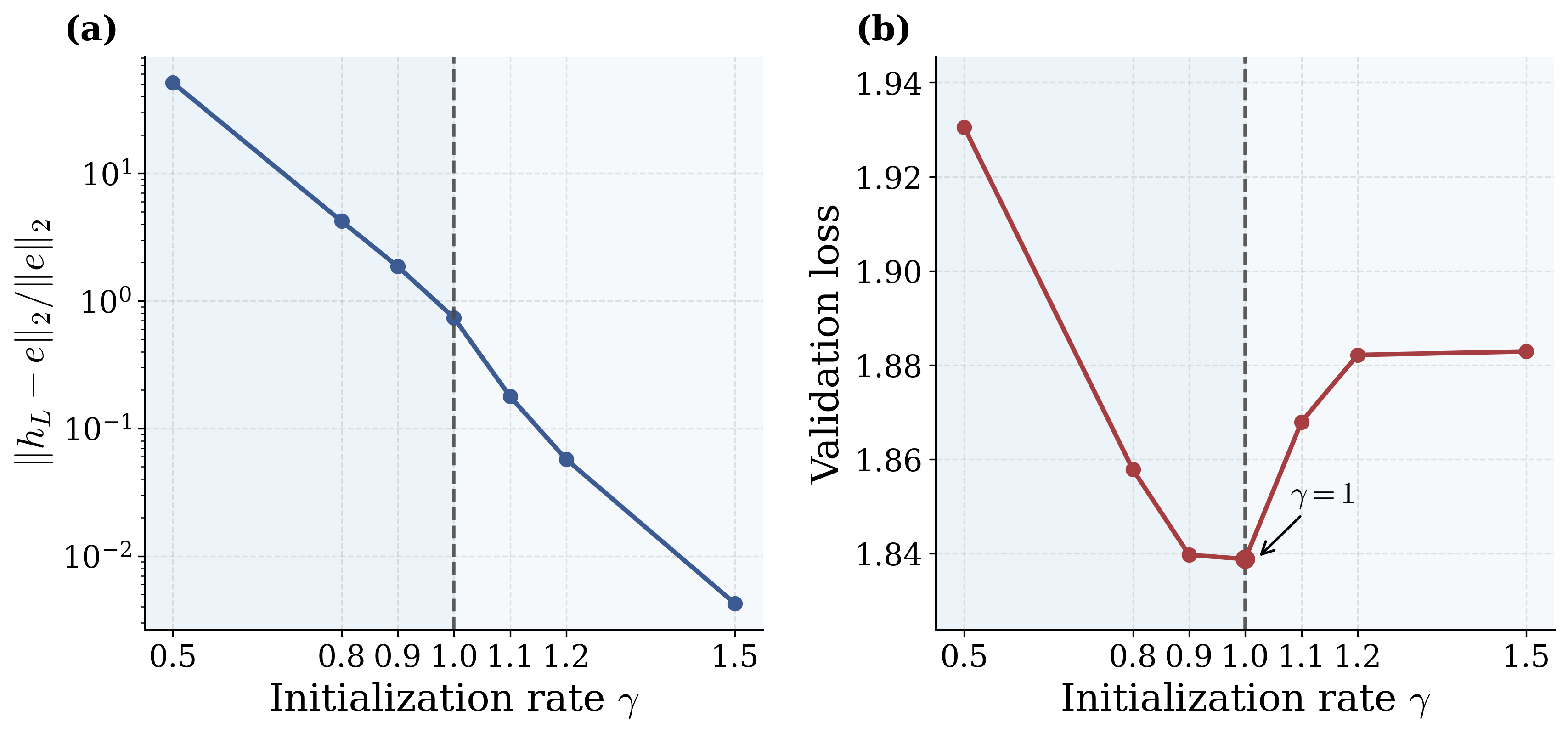}
    \caption{(a) Relative residual-flow strength $\|\vh_L-\ve\|_2/\|\ve\|_2$ as a function of initialization scale $\gamma$. (b) Final validation loss as a function of initialization scale $\gamma$.}
    \label{fig:gamma-loss}
\end{figure}

Both the scaling and its consequence are confirmed in experiments: the measured norm ratio follows $d^{1-\gamma}$ (Figure~\ref{fig:gamma-loss}a), and the pretraining loss improves from $\gamma=0.5$ to $\gamma=1$ but deteriorates beyond it (Figure~\ref{fig:gamma-loss}b). The optimal scale therefore sits at the balance point $\gamma=1$, small enough to induce condensation yet large enough to keep the network trainable.

\subsection{Where the gains come from: a token-level analysis}
\label{sec:token_level_analysis}

Validation loss and benchmark scores show that small initialization helps on average, but not where the improvement originates. We therefore ask whether the gain is spread uniformly across tokens or concentrated on specific ones. For each context $\vx_{<t}$ and label token $y_t$, we measure the symmetric probability gap between the small- and standard-initialization models,
\[\Delta^{\mathrm{sym}} p_t=\frac{2\left(p_{\mathrm{small}}(y_t \mid \vx_{<t})-p_{\mathrm{standard}}(y_t \mid \vx_{<t})\right)}{p_{\mathrm{small}}(y_t \mid \vx_{<t})+p_{\mathrm{standard}}(y_t \mid \vx_{<t})},\]
which is positive when small initialization assigns higher probability to the correct token.

\begin{figure}[htbp]
    \centering
    \includegraphics[width=0.9\textwidth]{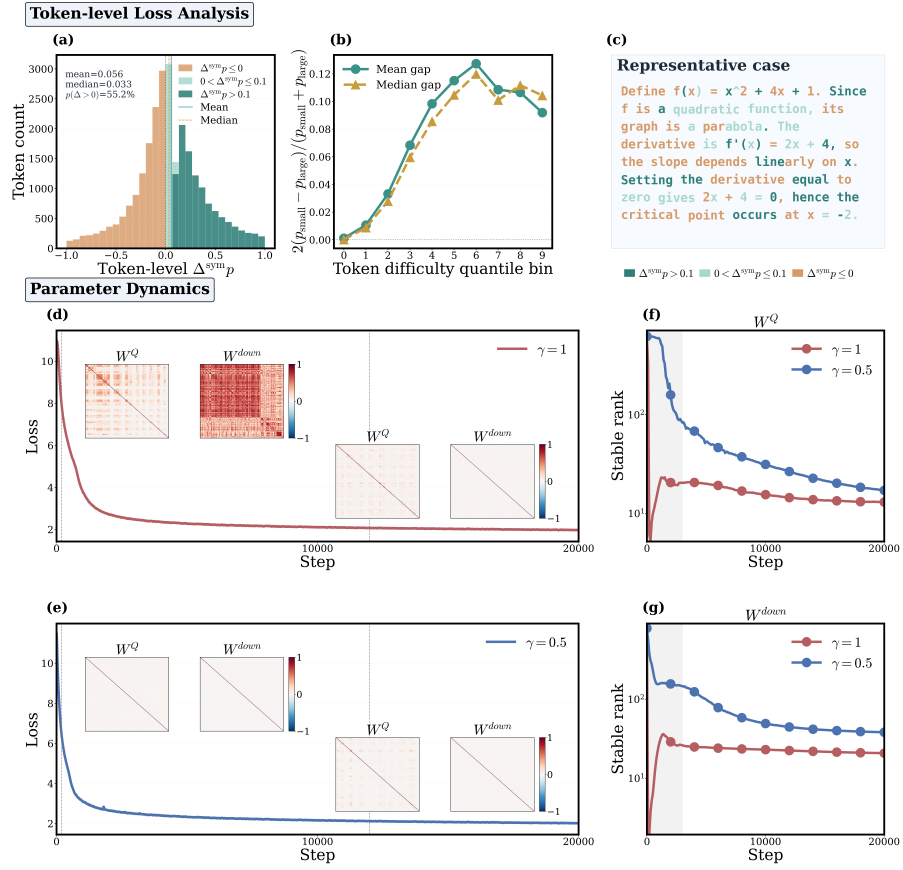}
    \caption{(a) Distribution of $\Delta^{\mathrm{sym}}p$ between $\gamma=1$ and $\gamma=0.5$. Vertical dashed lines indicate the mean and median values. (b) Mean and median $\Delta^{\mathrm{sym}}p$ across token difficulty quantile bins. (c) Representative token-level example with color-coded token-wise improvements according to the $\Delta^{\mathrm{sym}}p$. (d) Training loss curve and parameter cosine similarity snapshots for the $\gamma=1$ model. Insets visualize the first-layer $W^Q$ and $W^{down}$ at different training stages. (e) Training loss curve and parameter cosine similarity snapshots for the $\gamma=0.5$ model. Insets visualize the first-layer $W^Q$ and $W^{down}$ at different training stages. (f) Stable rank evolution of the first-layer $W^Q$ during training under different initialization scales. (g) Stable rank evolution of the first-layer $W^{down}$ during training under different initialization scales.}
    \label{fig:token_level_case_analysis}
\end{figure}

Over a validation set of 4B tokens, the distribution of $\Delta^{\mathrm{sym}} p$ is far from uniform (Figure~\ref{fig:token_level_case_analysis}a): most tokens change little or even slightly worsen, while a subset improves substantially. The aggregate loss reduction is thus driven by large gains on a minority of tokens rather than a uniform shift across the vocabulary.

To identify these tokens, we rank them by difficulty, defined as the average loss of the two models, $d_t=\tfrac{1}{2}\left(\ell_{\mathrm{small}}(y_t)+\ell_{\mathrm{std}}(y_t)\right)$, and bin them into ten equal-sized groups. The gain peaks on moderately-to-highly difficult tokens, not the hardest ones (Figure~\ref{fig:token_level_case_analysis}b): the hardest tokens are typically noisy fragments or rare symbols that neither model can predict, whereas intermediate-difficulty tokens demand contextual integration and reasoning, precisely where small initialization helps. A representative example confirms this (Figure~\ref{fig:token_level_case_analysis}c): in a short derivation about a quadratic function, the tokens that must be inferred from context, the derivative, the slope, the zero-gradient condition, and the resulting critical point, all show clear gains.

The benefit of small initialization is therefore not a uniform drop in loss but a sharpening of predictions that depend on earlier context. It strengthens the model's use of local dependencies and reasoning, and this systematic advantage on non-trivial yet learnable tokens accounts for the aggregate improvement.

\subsection{Mechanism: condensation drives a low-to-high complexity trajectory}

A large body of work has shown that, under small initialization, parameters within a layer first align into a few directions and later diversify, a pattern known as condensation~\cite{luo2021phase,zhou2022towards,chen2026from}. We ask whether LLMs follow the same path, tracking two measures of each parameter matrix during training: the row-wise cosine similarity and the stable rank, defined as $\frac{\|\vW\|_F^2}{\|\vW\|_2^2}=\frac{\sum_{i=1}^d\sigma_i^2}{\sigma_{\max}^2}$, where $\|\vW\|_F$ is the Frobenius norm and $\|\vW\|_2$ is the spectral norm, $\sigma_i$ denotes the singular value and $\sigma_{\max}$ means the maximum one. The stable rank is a continuous relaxation of the matrix rank. 

The two scales diverge sharply (Figure~\ref{fig:token_level_case_analysis}d,e). At initialization, the stable rank under two settings are both high due to the random initialization. Under $\gamma=1$, the similarity heatmaps reveal pronounced condensation early in training, with rows aligned into a few coherent blocks that gradually weaken as the matrix develops richer directions. Under $\gamma=0.5$, no such low-complexity phase appears: the structure is diffuse from the outset and evolves only slowly. The stable rank makes this quantitative (Figure~\ref{fig:token_level_case_analysis}f,g, Figure~\ref{fig:sr_f1} - \ref{fig:sr_f4}): under $\gamma=1$ it drops steeply at the start and then climbs, tracing a clear low-to-high complexity trajectory, whereas under $\gamma=0.5$ it begins high and declines slowly. It's noted that the high stable rank at initialization for both settings is expected, since the parameters are still random and random matrices typically have relatively high effective rank before training.

Small initialization therefore reshapes how representations form: the model first compresses into simple, low-rank structures and only later expands into more complex ones. This condensation-driven trajectory, rather than a mere change in final weights, distinguishes small initialization from the standard regime.

This low-to-high complexity trajectory also offers an explanation for the improved reasoning observed earlier. Because complexity grows only gradually, the model is driven to fit the data with the lowest complexity at each stage, seeking the fewest rules that account for the observations before resorting to more intricate ones. Such a bias toward minimal, parsimonious explanations favors genuine underlying reasoning over surface memorization, consistent with the larger token-level gains on predictions that require contextual inference.

\section{Discussion}

Our work establishes initialization scale as a meaningful design axis for LLMs training. We show that reducing the initialization scale can consistently improve pretraining loss and downstream performance in both dense and MoE LLMs. These results challenge the common view that initialization is merely a low-level implementation detail whose effect disappears during large-scale training.

A key implication is that the benefit of small initialization may be limited by the architecture with a larger model size. The reduced gain observed at larger scales does not reflect an intrinsic limitation of small initialization. Rather, it reveals a mismatch between the small initialization and standard LLMs components. Normalization constants can mask the intended scale difference, while attention sink can become more severe and limit the effect of small initialization. Correcting these factors releases the hidden gain of small initialization, showing that initialization and architecture should be considered jointly rather than independently.

Our results further show that the initialization scale must be chosen with a residual-flow balance in mind. Making the weights arbitrarily small is not beneficial: when \(\gamma>1\), residual updates become too weak relative to the embedding stream, and training degrades. The point \(\gamma=1\) therefore plays a special role, balancing reduced initial complexity with active residual updates. This provides a practical default for small-initialization training and a theoretical guide for tuning initialization scale.

We uncover that the benefits of small initialization stem from a change in the model's learning dynamics. Rather than producing a uniform improvement over all tokens, it mainly improves non-trivial, context-constrained predictions. At the parameter level, it induces a low-to-high complexity trajectory, where weight matrices first condense into simpler structures and later expand toward richer representations. These findings suggest that initialization affects the final loss via reshaping the formation process of internal representations.

We therefore advocate treating \(\gamma\)-controlled initialization as a native component of large language model design. Future training frameworks should expose \(\gamma\) as an explicit hyperparameter rather than hiding initialization scale inside default implementations. In particular, \(\gamma=1\) offers a principled default that combines small-scale regularization with residual-flow balance. More broadly, our findings suggest that effective LLM pretraining depends not only on scale, data, optimization, and architecture, but also on how representational complexity is initialized and allowed to emerge during learning.

\section{Method}

\subsection{Model Architecture}

\textbf{Dense model}: 
 We adopt a standard decoder-only Transformer architecture following the dense design and pre‑norm structure. The parameter scales are based on the four parameter scales of GPT‑3 \cite{brown2020language}, but we replace LayerNorm and MLP with the more commonly used RMSNorm and SwiGLU. Each Transformer block consists of the following components: RMSNorm, a multi‑head attention layer, and a feed‑forward network (MLP) with SwiGLU activation. Detailed parameters are shown in Table ~\ref{tab:dense_model_and_training_hyperparams}.

\textbf{MoE}: For the MoE architecture, we evaluated two attention mechanisms: Multi-head Latent Attention and conventional Multi-Head Attention. For the sparse expert layers, we followed the design practice by substantially increasing the number of experts while reducing the parameter scale of each individual expert, and by incorporating shared experts \cite{liu2024deepseek}. Detailed parameters are shown in Table ~\ref{tab:moe_model_and_training_hyperparams}.

\subsection{Initialization Scheme}
The main variable in our study is the initialization scale of model parameters. For each weight matrix $\vW$, we initialize its entries independently from a zero-mean Gaussian distribution:
\begin{equation*}
\vW_{i,j}\sim\mathcal{N}\left(0,d_{\mathrm{in}}^{-2\gamma}\right),
\end{equation*}
where $d_{\mathrm{in}}$ is the input dimension of $\vW$, and $\gamma$ controls the initialization scale. In particular, when $\gamma=\frac{1}{2}$, the variance becomes
$\mathrm{Var}(\vW_{i,j})=d_{\mathrm{in}}^{-1}$,
which corresponds to the conventional Xavier-like scaling. When $\gamma$ is increased, the weights are initialized with a smaller magnitude.

\subsection{Training Configuration}

We train all models using the standard next-token prediction objective. Training is conducted with Megatron-LM \cite{megatron-lm}, using AdamW \cite{loshchilov2017decoupled} as the optimizer. Unless otherwise specified, each model is trained for one epoch. The detailed training hyperparameters, are summarized in Tables ~\ref{tab:dense_model_and_training_hyperparams} and ~\ref{tab:moe_model_and_training_hyperparams}.

Training is performed with bfloat16 mixed precision, while gradient accumulation is conducted in fp32 precision. We disable both attention dropout and hidden dropout.

\subsection{Data}
The experiments were conducted on a high-quality bilingual (Chinese–English) corpus containing 1 trillion tokens and spanning multiple domains, including web data, mathematics, code, and books. The Dense model was trained on a 36B-token subset comprising 13.5B web tokens, 9B Wikipedia tokens, 4.5B code tokens, and 9B mathematics tokens. The MoE model was trained on 100B tokens sampled from the same corpus, with the original domain distribution preserved.

\subsection{Evaluation Protocol}
All downstream evaluations are conducted using the lm-evaluation-harness \cite{eval-harness}. The evaluation suite covers knowledge and question answering benchmarks, including ARC-C \cite{clark2018think}, TriviaQA \cite{2017arXivtriviaqa}, and MMLU \cite{hendrycks2021ethics,hendryckstest2021}; commonsense and general reasoning benchmarks, including HellaSwag \cite{zellers2019hellaswag}, BBH \cite{suzgun2023challenging}, and SocialIQA \cite{sap2019social}; and mathematical reasoning benchmarks, including GSM8K \cite{cobbe2021training} and MATH500 \cite{hendrycks2021measuring}. 
\subsection{Architectural Adjustments}

To examine whether the benefit of small initialization is limited by architectural components under larger scale, we consider two simple architectural adjustments: reducing the normalization constant $\varepsilon$ and introducing gated attention.

\paragraph{LayerNorm/RMSNorm constant}
Normalization layers rescale hidden states and can therefore interact with initialization scale.  For a hidden state $\vh\in\mathbb{R}^d$, LayerNorm is defined as
\begin{equation}
\mathrm{LN}(\vh)=\vgamma \odot\frac{\vh-\vmu(\vh)}{\sqrt{\sigma^2(\vh)+\varepsilon}}+\vbeta,
\end{equation}
where $\vmu(\vh)$ denotes the coordinate-wise mean, 
$\sigma^2(\vh)=d^{-1}\sum_i(\vh_i-\mu(\vh))^2$ is the hidden-state variance, 
$\varepsilon$ is the numerical constant for stability, and $\vgamma,\vbeta$ are learnable affine parameters. In our implementation, the Transformer blocks use RMSNorm, which is defined as
\begin{equation}
\mathrm{RMSNorm}(\vh)=\vgamma \odot\frac{\vh}{\sqrt{d^{-1}\sum_i \vh_i^2+\varepsilon}}.
\end{equation}
When the hidden-state variance becomes smaller than $\varepsilon$, the normalization factor becomes dominated by $\varepsilon$ rather than by the hidden-state scale. We compare two values of $\varepsilon$: the standard setting $\varepsilon=10^{-5}$ and a smaller setting $\varepsilon=10^{-12}$.

\paragraph{Gated attention}
We introduce gated attention to mitigate the attention sink phenomenon. 
Following recent work on gated softmax attention~\cite{qiu2026gated}, we apply a query-dependent gate to the output of each attention head. 
Given the standard attention heads $\mathrm{head}_1,\ldots,\mathrm{head}_{n_h}$, the gated attention output is written as
\begin{equation}
\mathrm{GAttn}(\vx)
=
\mathrm{Concat}
\left(
g_1(\vx)\odot \mathrm{head}_1,
\ldots,
g_{n_h}(\vx)\odot \mathrm{head}_{n_h}
\right)W_O,
\end{equation}
where $n_h$ is the number of attention heads, $W_O$ is the output projection matrix, and $g_h(\vx)$ is a head-specific sigmoid gate.

In practice, the gate is computed from the normalized hidden state before the attention module. 
For each head $h$, we use a linear projection followed by a sigmoid activation:
\begin{equation}
g_h(\vx)=\sigma(\vx W_h^g),
\end{equation}
where $W_h^g$ is the gate projection for head $h$, and $\sigma(\cdot)$ denotes the sigmoid function. 
The gate modulates the magnitude of the corresponding attention head after the softmax attention weights have been computed. 
Thus, the gating mechanism does not change the causal attention mask or the softmax normalization itself; it only controls how strongly each head contributes to the residual update.

\subsection{Mechanistic Analysis Metrics}

\paragraph{Attention sink score}
For layer $\ell$ and head $h$, let $A_h^{(\ell)}(t,i,j)$ denote the attention weight from query position $i$ to key position $j$ in sequence $t$. 
We define the sink score as
\begin{equation}\label{eq:sink_score}
S_{\mathrm{sink}}^{(\ell,h)}(t)=\frac{1}{T}\sum_{i=1}^{T}A_h^{(\ell)}(t,i,1).
\end{equation}
This quantity measures the average attention mass assigned to the first token by all query positions in a sequence. 
A larger $S_{\mathrm{sink}}^{(\ell,h)}(t)$ indicates stronger attention concentration on the first token. 
In our experiments, we report layer-wise sink scores by averaging $S_{\mathrm{sink}}^{(\ell,h)}(t)$ over validation sequences and attention heads.
\paragraph{Token-level loss analysis}
To analyze where the gain of small initialization comes from, we compare the prediction behavior of the small-initialization model and the standard-initialization model at the token level. For each validation token $y_t$, we denote the probability assigned to the ground-truth token by the small-initialization model and the standard-initialization model as
\[
p_{\mathrm{small}}(y_t \mid x_{<t})\quad \text{and} \quad p_{\mathrm{std}}(y_t \mid x_{<t}),
\]
respectively. The corresponding token-level losses are defined as
\[
\ell_{\mathrm{small}}(y_t)=-\log p_{\mathrm{small}}(y_t \mid x_{<t}),\qquad \ell_{\mathrm{std}}(y_t)=-\log p_{\mathrm{std}}(y_t \mid x_{<t}).
\]

A direct comparison of token-level losses shows whether small initialization reduces the cross-entropy loss on each token. However, raw loss differences can be strongly affected by the intrinsic difficulty of individual tokens. Therefore, in addition to token-level loss, we analyze the probability assigned to the correct token.

We first define the absolute correct-token probability gap as
\[
\Delta p_t=p_{\mathrm{small}}(y_t \mid x_{<t})-p_{\mathrm{std}}(y_t \mid x_{<t}).
\]
To normalize the scale of the probability difference, we further use the symmetric probability gap:
\[
\Delta^{\mathrm{sym}} p_t=\frac{2\left(p_{\mathrm{small}}(y_t \mid x_{<t})-p_{\mathrm{std}}(y_t \mid x_{<t})\right)}{p_{\mathrm{small}}(y_t \mid x_{<t})+p_{\mathrm{std}}(y_t \mid x_{<t})}.
\]
This metric measures the relative improvement in the correct-token probability. It is positive when the small-initialization model assigns a higher probability to the ground-truth token, and negative when the standard-initialization model assigns a higher probability.

To examine how token-level improvement depends on prediction difficulty, we define the difficulty of each token as the average token loss of the two models:
\[
d_t=\frac{1}{2}\left(\ell_{\mathrm{small}}(y_t)+\ell_{\mathrm{std}}(y_t)\right).
\]
We sort all validation tokens according to $d_t$ and divide them into ten equal-sized difficulty bins. For each bin, we compute the mean and median values of $\Delta^{\mathrm{sym}} p_t$.

\paragraph{Cosine similarity and stable rank}
To characterize the training dynamics induced by different initialization scales, we analyze the structure of model parameters during training. We focus on two complementary measurements: row-wise cosine similarity and stable rank.

For a weight matrix $\vW\in\mathbb{R}^{m\times n}$, let $\vw_i\in\mathbb{R}^n$ denote its $i$-th row vector. 
We compute the pairwise cosine similarity between rows as
\begin{equation}
C_{ij}(\vW)
=
\frac{\langle \vw_i,\vw_j\rangle}
{\|\vw_i\|_2\|\vw_j\|_2},
\end{equation}
where $C_{ij}(\vW)\in[-1,1]$. 
The resulting matrix $C(\vW)\in\mathbb{R}^{m\times m}$ describes the angular similarity among the row vectors of $\vW$. 

To quantify the effective dimensionality of $\vW$, we further compute its stable rank, defined as
\begin{equation}
\mathrm{StableRank}(\vW)=\frac{\|\vW\|_F^2}{\|\vW\|_2^2}=\frac{\sum_{i=1}^d\sigma_i^2}{\sigma_{\max}^2},
\end{equation}
where $\sigma_i$ denotes the singular value and $\sigma_{\max}$ means the maximum one. 
The stable rank is a continuous relaxation of the matrix rank. 


\bmhead{Acknowledgements}
This work is sponsored by the National Key R$\&$D Program
of China Grant No. 2022YFA1008200  (Z. X.), the National Natural Science Foundation of China Grant No. 92570001 (Z. X.), 12371511 (Z. X.), 12422119 (Z. X.), 2025 Key Technology R\&D Program ``New Generation Information Technology'' Project 25511103100 of Shanghai Municipal Science and Technology Commission (Z. X.).

\begin{appendices}

\section{Detailed analysis}\label{app:proof}

We analyze the relative scale of the accumulated residual update $\vh_L-\ve$ compared with the initial embedding stream $\ve$. Consider a residual network of the form
\begin{equation*}
\vh_0=\ve,\qquad \vh_{\ell+1}=\vh_\ell+\vF_\ell(\vh_\ell),\qquad \ell=0,\dots,L-1.
\end{equation*}
Unrolling the residual recursion gives
\begin{equation*}
\vh_L=\ve+\sum_{\ell=0}^{L-1}\vF_\ell(\vh_\ell),
\end{equation*}
and therefore
\begin{equation*}
\vh_L-\ve=\sum_{\ell=0}^{L-1}\vF_\ell(\vh_\ell).
\end{equation*}
Under small initialization, the residual branches are initially weak. Thus, in the leading-order scale analysis, the output of each block can be approximated by applying the corresponding residual branch to the initial embedding stream:
\begin{equation*}
\vF_\ell(\vh_\ell)\approx \vF_\ell(\ve).
\end{equation*}
Hence,
\begin{equation*}
\vh_L-\ve\approx \sum_{\ell=0}^{L-1}\vF_\ell(\ve).
\end{equation*}

We assume that the embedding vector and the weight matrices are initialized as
\begin{equation*}
\ve_i\sim\mathcal{N}(0,d^{-2\gamma}),\qquad \vW_{ij}\sim\mathcal{N}(0,d^{-2\gamma}).
\end{equation*}
Then the squared norm of the embedding satisfies
\begin{equation*}
\mathbb{E}\lVert \ve\rVert_2^2=\sum_{i=1}^{d}\mathbb{E}\ve_i^2=d\cdot d^{-2\gamma}=d^{1-2\gamma}.
\end{equation*}
Thus the typical scale of the embedding norm is
\begin{equation*}
\lVert \ve\rVert_2=\Theta(d^{1/2-\gamma}).
\end{equation*}

Now we compute the scale of one residual module applied to $\ve$. We take the module to be an RMSNorm followed by two linear transformations:
\begin{equation*}
\vF_\ell(\ve)=\vW_{2,\ell}\phi\big(\vW_{1,\ell}\operatorname{RMSNorm}(\ve)\big).
\end{equation*}
Ignoring the constant affine weight and assuming that the RMS term dominates the numerical $\varepsilon$, RMSNorm can be written as
\begin{equation*}
\operatorname{RMSNorm}(\ve)=\frac{\ve}{\operatorname{RMS}(\ve)}.
\end{equation*}
Here
\begin{equation*}
\operatorname{RMS}(\ve)=\left(\frac{1}{d}\sum_{i=1}^{d}e_i^2\right)^{1/2}=\frac{\lVert \ve\rVert_2}{\sqrt{d}}.
\end{equation*}
Since $\lVert \ve\rVert_2=\Theta(d^{1/2-\gamma})$, we have
\begin{equation*}
\operatorname{RMS}(\ve)=\Theta(d^{-\gamma}).
\end{equation*}
Therefore,
\begin{equation*}
\lVert \operatorname{RMSNorm}(\ve)\rVert_2=\left\lVert \frac{\ve}{\operatorname{RMS}(\ve)}\right\rVert_2=\Theta\left(\frac{d^{1/2-\gamma}}{d^{-\gamma}}\right)=\Theta(d^{1/2}).
\end{equation*}
Thus RMSNorm removes the initialization scale of $\ve$ and maps it to a vector with coordinate scale $O(1)$ and Euclidean norm $\Theta(d^{1/2})$.

For a random matrix $\vW\in\mathbb{R}^{d\times d}$ with entries initialized as $\mathcal{N}(0,d^{-2\gamma})$, its spectral norm has the typical scale
\begin{equation*}
\lVert \vW\rVert_2=\Theta(d^{1/2-\gamma}).
\end{equation*}
Therefore, after the first linear transformation,
\begin{equation*}
\left\lVert \vW_{1,\ell}\operatorname{RMSNorm}(\ve)\right\rVert_2=\Theta(d^{1/2-\gamma})\cdot \Theta(d^{1/2})=\Theta(d^{1-\gamma}).
\end{equation*}
We assume that the activation function in the FFN block contributes only an $O(1)$ factor to the leading-order scale, so that
\begin{equation*}
\left\lVert \phi\big(\vW_{1,\ell}\operatorname{RMSNorm}(\ve)\big)\right\rVert_2=\Theta(d^{1-\gamma}).
\end{equation*}
Applying the second linear transformation gives
\begin{equation*}
\lVert \vF_\ell(\ve)\rVert_2=\left\lVert \vW_{2,\ell}\phi\big(\vW_{1,\ell}\operatorname{RMSNorm}(\ve)\big)\right\rVert_2.
\end{equation*}
Using again $\lVert \vW_{2,\ell}\rVert_2=\Theta(d^{1/2-\gamma})$, we obtain
\begin{equation*}
\lVert \vF_\ell(\ve)\rVert_2=\Theta(d^{1/2-\gamma})\cdot \Theta(d^{1-\gamma})=\Theta(d^{3/2-2\gamma}).
\end{equation*}

Therefore, each residual module output has scale
\begin{equation*}
\lVert \vF_\ell(\ve)\rVert_2=\Theta(d^{3/2-2\gamma}).
\end{equation*}
Using
\begin{equation*}
\vh_L-\ve\approx \sum_{\ell=0}^{L-1}\vF_\ell(\ve),
\end{equation*}
and treating $L$ as fixed with respect to $d$, the accumulation over layers does not change the leading exponent in $d$. Thus,
\begin{equation*}
\lVert \vh_L-\ve\rVert_2=\Theta(d^{3/2-2\gamma}).
\end{equation*}

Finally, comparing this with the initial embedding scale
\begin{equation*}
\lVert \ve\rVert_2=\Theta(d^{1/2-\gamma}),
\end{equation*}
we obtain
\begin{equation*}
\frac{\lVert \vh_L-\ve\rVert_2}{\lVert \ve\rVert_2}=\Theta\left(\frac{d^{3/2-2\gamma}}{d^{1/2-\gamma}}\right)=\Theta(d^{1-\gamma}).
\end{equation*}
Therefore, the relative scale between the accumulated residual update and the initial embedding stream is
\begin{equation*}
\frac{\lVert \vh_L-\ve\rVert_2}{\lVert \ve\rVert_2}\sim d^{1-\gamma}.
\end{equation*}

This gives three regimes. When $\gamma<1$,
\begin{equation*}
\frac{\lVert \vh_L-\ve\rVert_2}{\lVert \ve\rVert_2}\to \infty,
\end{equation*}
so the residual update becomes larger than the embedding stream. When $\gamma=1$,
\begin{equation*}
\frac{\lVert \vh_L-\ve\rVert_2}{\lVert \ve\rVert_2}=\Theta(1),
\end{equation*}
so the residual update and the embedding stream remain comparable. When $\gamma>1$,
\begin{equation*}
\frac{\lVert \vh_L-\ve\rVert_2}{\lVert \ve\rVert_2}\to 0,
\end{equation*}
so the residual update becomes smaller than the embedding stream.

Thus, under the small-initialization approximation and for a pre-normalized FFN block with two linear transformations, $\gamma=1$ is the balance point at which the accumulated residual update $\vh_L-\ve$ and the initial embedding stream $\ve$ have the same leading-order scale.

\section{Model Architecture}

\begin{table*}[htbp]
\centering
\caption{Architecture and training hyperparameters of the dense decoder-only Transformer models.}
\label{tab:dense_model_and_training_hyperparams}
\small
\setlength{\tabcolsep}{5pt}
\begin{tabular}{lcccc}
\toprule
\textbf{Hyperparameter}
& \textbf{0.1B}
& \textbf{0.3B}
& \textbf{0.7B}
& \textbf{1.5B} \\
\midrule

\multicolumn{5}{l}{\textbf{Architecture configuration}} \\
\midrule
Architecture type
& \multicolumn{4}{c}{Decoder-only Transformer} \\
Design type
& \multicolumn{4}{c}{Dense} \\
Normalization scheme
& \multicolumn{4}{c}{Pre-norm} \\
Attention mechanism
& \multicolumn{4}{c}{Multi-head attention} \\
Activation function
& \multicolumn{4}{c}{SwiGLU} \\
Normalization type
& \multicolumn{4}{c}{RMSNorm} \\
Positional encoding
& \multicolumn{4}{c}{Rotary Position Embedding} \\
Vocabulary size, $|\mathcal{V}|$
& \multicolumn{4}{c}{60416} \\

\midrule
Number of layers, $N_{\mathrm{layer}}$
& 12 & 24 & 24 & 24 \\
Hidden size, $d_{\mathrm{model}}$
& 768 & 1024 & 1024 & 2048 \\
Number of attention heads, $N_{\mathrm{head}}$
& 12 & 16 & 16 & 24 \\
Attention head dimension, $d_{\mathrm{head}}$
& 64 & 64 & 96 & 128 \\
MLP intermediate size, $d_{\mathrm{ff}}$
& 2304 & 3072 & 4096 & 5460 \\
Non-embedding parameters
& $\sim$92M & $\sim$327M & $\sim$680M & $\sim$1.5B \\
Gated-attention parameters
& $\sim$7M & $\sim$25M & $\sim$56M & $\sim$150M \\

\midrule
\multicolumn{5}{l}{\textbf{Training configuration}} \\
\midrule
Learning-rate ratio, $r_{\mathrm{lr}}$
& \multicolumn{4}{c}{0.1} \\
Warm-up ratio, $r_{\mathrm{warmup}}$
& \multicolumn{4}{c}{0.03} \\
Peak learning rate, $\eta_{\mathrm{peak}}$
& 6e-4 & 3e-4 & 2.5e-4 & 2e-4 \\
Batch size, $B$
& 256 & 256 & 256 & 512 \\

\bottomrule
\end{tabular}
\end{table*}

\clearpage

\begin{table*}[htbp]
\centering
\caption{Architecture and training hyperparameters of the MoE decoder-only Transformer models.}
\label{tab:moe_model_and_training_hyperparams}
\small
\setlength{\tabcolsep}{6pt}
\begin{tabular}{lcc}
\toprule
\textbf{Hyperparameter}
& \textbf{1.5B-A250M}
& \textbf{3.2B-A0.5B} \\
\midrule

\multicolumn{3}{l}{\textbf{Architecture configuration}} \\
\midrule
Architecture type
& \multicolumn{2}{c}{Decoder-only Transformer} \\
Design type
& \multicolumn{2}{c}{Sparse MoE} \\
Normalization scheme
& \multicolumn{2}{c}{Pre-norm} \\
Attention mechanism
& \multicolumn{2}{c}{Multi-head attention} \\
Activation function
& \multicolumn{2}{c}{SwiGLU} \\
Normalization type
& \multicolumn{2}{c}{RMSNorm} \\
Positional encoding
& \multicolumn{2}{c}{Rotary Position Embedding} \\
Vocabulary size, $|\mathcal{V}|$
& \multicolumn{2}{c}{151643} \\
Maximum sequence length, $L_{\mathrm{ctx}}$
& \multicolumn{2}{c}{4096} \\

\midrule
\multicolumn{3}{l}{\textbf{Backbone configuration}} \\
\midrule
Number of dense layers, $N_{\mathrm{dense}}$
& 1 & 1 \\
Number of MoE layers, $N_{\mathrm{MoE}}$
& 11 & 18 \\
Hidden size, $d_{\mathrm{model}}$
& 512 & 768 \\
Number of attention heads, $N_{\mathrm{head}}$
& 8 & 12 \\
Number of KV heads, $N_{\mathrm{kv}}$
& 4 & 4 \\
Attention head dimension, $d_{\mathrm{head}}$
& 64 & 64 \\
Dense intermediate size, $d_{\mathrm{ff,dense}}$
& 10944 & 10944 \\

\midrule
\multicolumn{3}{l}{\textbf{MoE configuration}} \\
\midrule
MoE layer interval
& 1 & 1 \\
Number of routed experts per layer, $N_{\mathrm{expert}}$
& 145 & 145 \\
Number of activated experts, $K$
& 9 & 9 \\
Number of shared experts, $N_{\mathrm{shared}}$
& 1 & 1 \\
Expert intermediate size, $d_{\mathrm{expert}}$
& 512 & 512 \\
Shared expert intermediate size, $d_{\mathrm{shared}}$
& 512 & 512 \\
Router type
& \multicolumn{2}{c}{Top-$K$ routing} \\

\midrule
\multicolumn{3}{l}{\textbf{Training configuration}} \\
\midrule
Learning-rate ratio, $r_{\mathrm{lr}}$
& \multicolumn{2}{c}{0.1} \\
Warm-up ratio, $r_{\mathrm{warmup}}$
& \multicolumn{2}{c}{0.03} \\
Peak learning rate, $\eta_{\mathrm{peak}}$
& \multicolumn{2}{c}{4.2e-4} \\
Batch size, $B$
& \multicolumn{2}{c}{2048} \\

\midrule
\multicolumn{3}{l}{\textbf{Parameter scale}} \\
\midrule
Total parameters
& $\sim$ 1.4B & $\sim$ 3.2B \\
Activated parameters
& $\sim$ 260M & $\sim$ 470M \\

\bottomrule
\end{tabular}
\end{table*}


\section{Complete results of stable rank}
\begin{figure}[H]
    \centering
    \includegraphics[width=0.95\linewidth]{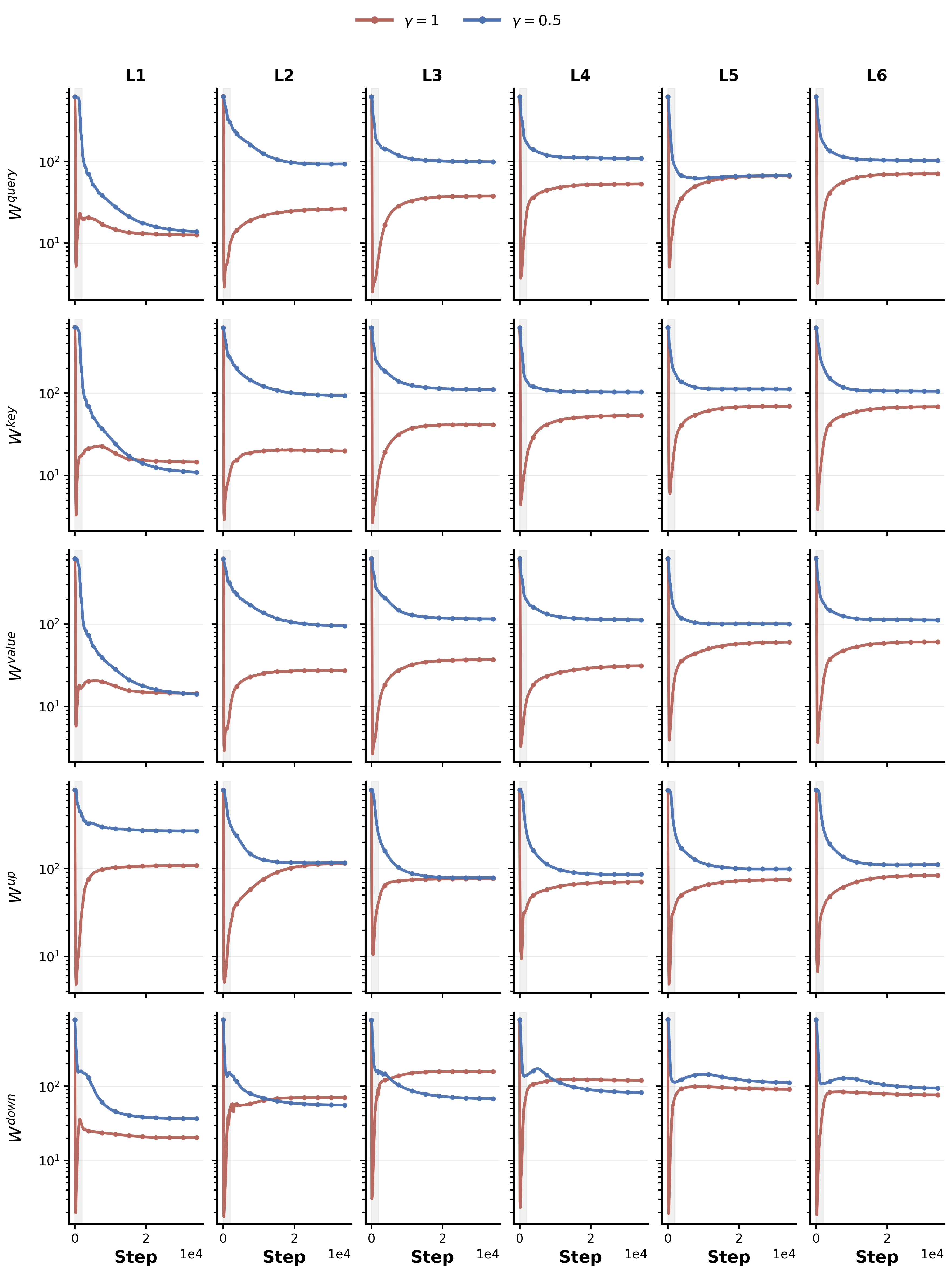}
    \caption{Stable-rank dynamics of linear modules in layers 1--6. Rows correspond to different matrix types, including query, key, value, FFN up-projection, and FFN down-projection matrices, while columns correspond to layers. Each subplot compares the stable-rank trajectory of the same matrix under $\gamma=0.5$ and $\gamma=1$, with the vertical axis shown on a logarithmic scale.}
    \label{fig:sr_f1}
\end{figure}

\clearpage

\begin{figure}[p]
    \centering
    \includegraphics[width=1\linewidth]{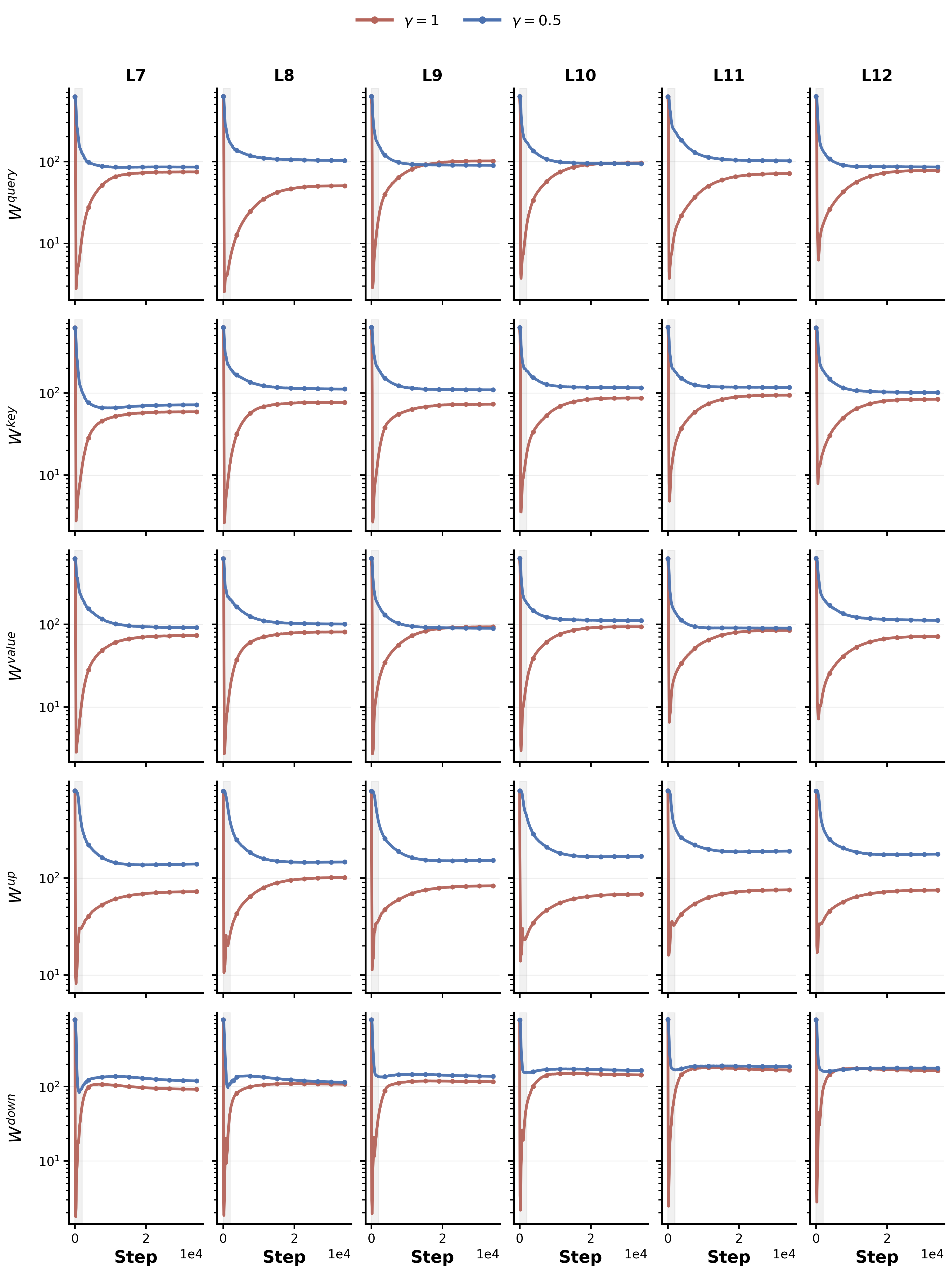}
    \caption{Stable-rank dynamics of linear modules in layers 7--12. Rows correspond to different matrix types, including query, key, value, FFN up-projection, and FFN down-projection matrices, while columns correspond to layers. Each subplot compares the stable-rank trajectory of the same matrix under $\gamma=0.5$ and $\gamma=1$, with the vertical axis shown on a logarithmic scale.}
    \label{fig:sr_f2}
\end{figure}

\clearpage

\begin{figure}[p]
    \centering
    \includegraphics[width=1\linewidth]{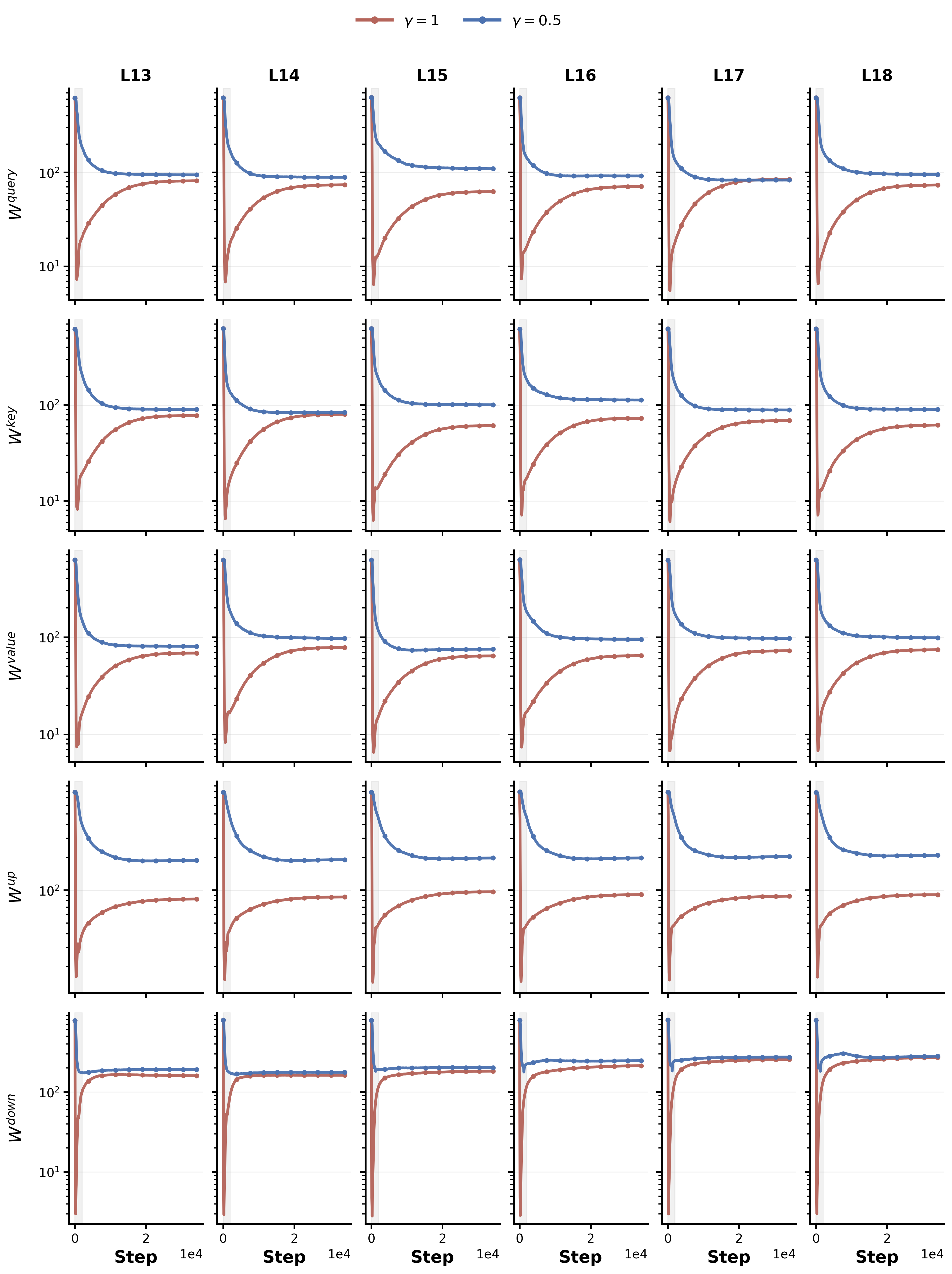}
    \caption{Stable-rank dynamics of linear modules in layers 13--18. Rows correspond to different matrix types, including query, key, value, FFN up-projection, and FFN down-projection matrices, while columns correspond to layers. Each subplot compares the stable-rank trajectory of the same matrix under $\gamma=0.5$ and $\gamma=1$, with the vertical axis shown on a logarithmic scale.}
    \label{fig:sr_f3}
\end{figure}

\clearpage

\begin{figure}[p]
    \centering
    \includegraphics[width=1\linewidth]{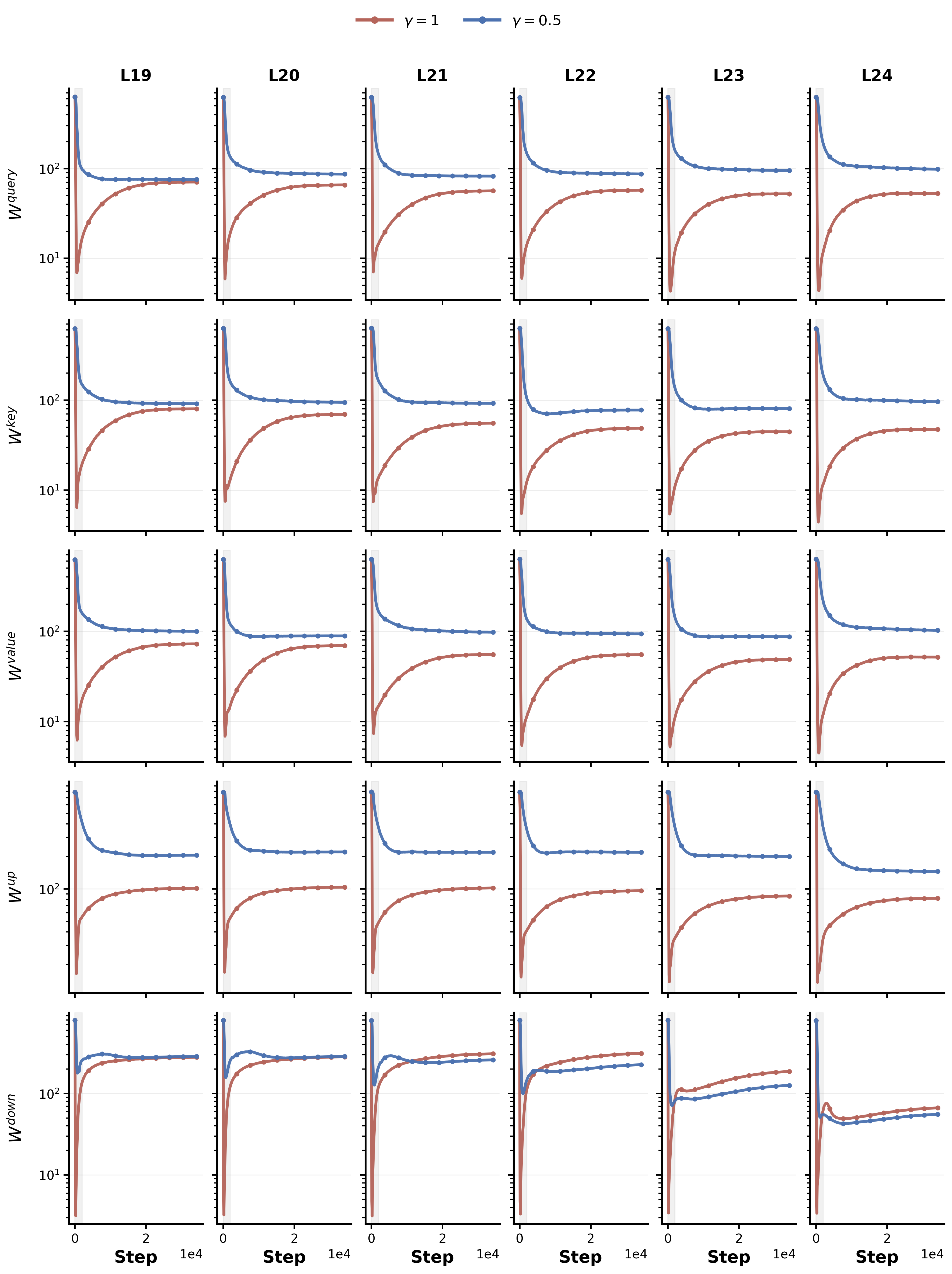}
    \caption{Stable-rank dynamics of linear modules in layers 19--24. Rows correspond to different matrix types, including query, key, value, FFN up-projection, and FFN down-projection matrices, while columns correspond to layers. Each subplot compares the stable-rank trajectory of the same matrix under $\gamma=0.5$ and $\gamma=1$, with the vertical axis shown on a logarithmic scale.}
    \label{fig:sr_f4}
\end{figure}

\end{appendices}

\clearpage

\bibliography{ref}

\end{document}